\documentclass[cameraready]{Interspeech}
\title{Iterative LLM-based improvement for French Clinical Interview Transcription and Speaker Diarization}

\author[affiliation={1,2}, orcid=0009-0009-4322-9108, correspondingauthor]{Ambre}{Marie}
\author[affiliation={3}, orcid=0009-0000-9994-0737]{Thomas}{Bertin}
\author[affiliation={1}, orcid=0000-0002-6958-4994]{Guillaume}{Dardenne}
\author[affiliation={1}, orcid=0000-0003-1669-7140]{Gwenolé}{Quellec}
\address{
    $^1$ LaTIM UMR 1101 INSERM, Brest, France \\
    $^2$ University of Western Brittany, Brest, France \\
    $^3$ University of Rouen Normandy, Rouen, France
}

\email{ambre.marie@univ-brest.fr, thomas.bertin@univ-rouen.fr, guillaume.dardenne@inserm.fr, gwenole.quellec@inserm.fr}


\usepackage{comment}

\begin{document}

\maketitle

\begin{abstract}
    Automatic speech recognition for French medical conversations remains challenging, with word error rates often exceeding 30\% in spontaneous clinical speech. This study proposes a multi-pass LLM post-processing architecture alternating between Speaker Recognition and Word Recognition passes to improve transcription accuracy and speaker attribution. Ablation studies on two French clinical datasets (suicide prevention telephone counseling and preoperative awake neurosurgery consultations) investigate four design choices: model selection, prompting strategy, pass ordering, and iteration depth. Using Qwen3-Next-80B, Wilcoxon signed-rank tests confirm significant WDER reductions on suicide prevention conversations (p $<$ 0.05, n=18), while maintaining stability on awake neurosurgery consultations (n=10), with zero output failures and acceptable computational cost (RTF 0.32), suggesting feasibility for offline clinical deployment, pending validation on larger corpora.

\end{abstract}

\section{Introduction}

Automatic speech recognition (ASR) still presents major limitations for concrete medical applications in French, where transcription errors can directly affect clinical analysis and downstream uses \cite{bertin2025transcription}. In clinical interviews, ASR performance remains far below that observed in controlled dictation settings. While dictation tasks often achieve Word Error Rates (WER) below 10\%, general-purpose ASR systems typically reach WERs between 30\% and 65\% on spontaneous clinical interviews. Even domain-trained or medically adapted systems only reduce these error rates to approximately 18-35\%, often failing to capture clinically important details under real-world conditions \cite{VanBuchem2021The,Renato2024Development}.

Medical conversations in non-English languages remain substantially harder for ASR than English dictation, mainly due to the limited availability of in-domain training data and language-specific challenges \cite{linke2025context}. In French, rich morphology and frequent homophones make lexical disambiguation particularly difficult: even state-of-the-art systems systematically confuse common homophones in conversational speech \cite{Pasandi2022Evaluation}. Moreover, high-quality labeled medical speech data in French and other non-English languages is much rarer than in English, especially for real clinical interviews rather than read or scripted speech \cite{Reitmaier2022Opportunities}.

Our baseline system (WhisperX large-v3 + Pyannote 3.1) achieves WDER of approximately 43\% on both datasets. This high error rate reflects both ASR limitations: lexical errors on medical terminology, and speaker diarization errors during rapid turn-taking and overlaps. The strict, linguist-produced reference transcriptions also preserves fine-grained phenomena of spontaneous speech production that are often suppressed by conventional ASR normalization pipelines, such as filled pauses, partial lexical initiations with restart, repetitions, and self-repairs.

Recent work has explored the use of large language models (LLMs) for post-processing ASR outputs. Adedeji et al. \cite{adedeji2024sound} investigated LLM-based post-processing for English primary care consultations using commercial models such as GPT-4, Claude, and Gemini, reporting reductions in WER as well as improvements in speaker-specific transcription quality. However, diarization improvements were mainly assessed qualitatively, and the study was limited to English data.

These limitations motivate the need for systematic evaluation of LLM post-processing approaches in non-English medical contexts. 
This work proposes a multi-pass LLM post-processing architecture using Qwen3-Next-80B (80B parameters) for French clinical interview transcriptions. The method iteratively alternates between Speaker Recognition (SR) and Word Recognition (WR) passes, and is validated in two distinct clinical domains: suicide prevention telephone counselling and preoperative awake neurosurgery consultations. Ablation studies demonstrate that this configuration achieves significant improvements in speaker attribution in our medical conversations while maintaining stability over two separate datasets.
Given the rarity and sensitivity of such recordings, this study is intentionally framed as a feasibility study rather than a definitive performance benchmark.

\section{Material and Methods}

The complete source code, including the transcription pipeline, LLM post-processing implementation, and domain-adapted prompts, is publicly available at \url{https://github.com/amarie-research/iterative-llm-clinical-transcription}.

\subsection{Datasets}
Experiments are conducted on two French medical speech datasets drawn from distinct clinical contexts. Both datasets consist of non-scripted, real-world interactions and exhibit the variability typical of routine medical conversations.
All recordings are in French and were manually transcribed by a linguist specialized in verbal interactions following the ICOR transcription convention \cite{groupe2013convention}. The transcriptions include fine-grained annotations of hesitations, pauses, overlapping speech, and speaker changes, providing a strict reference for evaluating both transcription and diarization performance. Consistent with French interactional linguistics conventions \cite{bertin2025transcription}, reference transcriptions omit capitalization and punctuation, replacing them with prosodic markers (micropauses, intonation contours).

Table~\ref{tab:dataset_stats} summarizes the scale and structure of the two datasets.

\subsubsection{Suicide Prevention (SP) dataset}
The Suicide Prevention (SP) dataset consists of telephone calls conducted as part of post-suicide-attempt follow-up. Each call begins with an automatic answering system, followed by an interview between a psychiatric nurse and a patient. In some cases, a second psychiatric nurse also participates in the exchange. Speech is spontaneous and conversational, with variable turn lengths, and high lexical diversity (reflecting the diversity of the life stories told). These conversations are explicitly focused on emotional assessment and therapeutic support, and the protocol prioritizes empathetic listening and patient elaboration.

\subsubsection{Awake Neurosurgery (AN) dataset}
This dataset includes audio recordings collected in a hospital consultation room during preoperative preparation for language testing in awake neurosurgery (AN). Conversations always involve a patient and at least one medical professional, such as a neurosurgeon, neuropsychologist, psychologist, or anaesthesiologist. Additional participants, including a nurse or a patient's family member, may also be present. These interviews follow a format with clearer turn-taking organization than SP. When excluding SP's automated voice system, AN involves more speakers on average, and exhibits higher conversational density with shorter median turn duration and more frequent very short turns, reflecting the structured question-answer dynamic typical of in person clinical assessments.
The medical context being about brain tumor resection, the consultation is explanatory while the conversation objective is cognitive and linguistic assessment, following a standardized neuropsychological protocol.

\begin{table}[ht!]
\centering
\caption{Dataset statistics}
\label{tab:dataset_stats}
\begin{tabular}{lcc}
\hline
 & AN & SP \\
\hline
\multicolumn{3}{l}{\textit{Dataset size}} \\
Total duration & 5h49m & 6h32m \\
Total words & 57,668 & 66,144 \\
Conversations & 10 & 18 \\
Unique speakers (corpus) & 15 & 13 \\
\hline
\multicolumn{3}{l}{\textit{Conversational structure}} \\
Speakers per conversation & 2--4 & 2--3 *\\
Avg. duration (min) & 35.0 ± 12.7 & 21.8 ± 5.1 \\
Segments per minute & 16.9 ± 4.2 & 14.7 ± 2.6 \\
Median turn duration (s) & 1.6 ± 0.6 & 2.2 ± 0.5 \\
Type-token ratio & 0.30 ± 0.07 & 0.49 ± 0.06 \\
\hline
\end{tabular}
\\AN: Awake Neurosurgery, SP: Suicide Prevention
\\ * does not include the automatic answering system
\end{table}

\subsection{Pipeline}

The initial transcription is performed using the WhisperX framework \cite{bain2022whisperx}. The large-v3 model is used, which offers a robust balance between accuracy and computational cost. To ensure precise alignment of transcription and diarization, the audio is first processed with a VAD (Voice Activity Detection) filter followed by the Whisper model to generate initial segments. Word-level timestamps are obtained using the VoxPopuli French ASR model \cite{wang2021voxpopuli} as forced alignment backbone within WhisperX. Speaker identification is performed using the Pyannote 3.1 architecture \cite{bredin2020pyannote}. This step assigns generic labels (e.g., SPEAKER\_00, SPEAKER\_01) to each segment.
This output, obtained without any LLM-based correction or post-processing, is referred to as the Baseline during the rest of this paper.

This Baseline serves as the input for the LLM post-processing stage. In this context, a segment is defined as a structured unit containing:
\begin{itemize}
    \item Timestamps: The start and end times of the utterance.
    \item Speaker Label: The identity assigned by the diarization model.
    \item Textual Content: The transcribed speech corresponding to that specific turn.
\end{itemize}
To handle large-scale clinical interviews, a chunking strategy has been implemented. Files exceeding 500 segments are split into sequential chunks of 500 segments to fit the LLM's context limits. Each chunk is processed independently, and the results are concatenated in order.
The pipeline is designed to be model-agnostic and supports both commercial APIs and open-source models. 

All experiments were conducted on an NVIDIA RTX A5000 GPU (24 GB VRAM) with 64 GB system RAM. Real-Time Factor (RTF) measurements (Section \ref{sec:rtf}) include only LLM inference time and exclude initial WhisperX/Pyannote processing (baseline RTF 0.09), allowing direct evaluation of post-processing computational overhead.

For evaluation, generic speaker labels produced by the baseline (SPEAKER\_00, SPEAKER\_01, etc.) are mapped to reference speaker roles using the Hungarian algorithm, which finds the optimal bijective assignment minimizing total word-level error across all speaker pairs, ensuring fair comparison with LLM post-processed outputs that produce named role labels.

\subsection{Proposed Method: N-pass Architecture}

The proposed pipeline consists of sequential LLM processing passes using the Qwen3-Next-80B model (see Section \ref{sec:model_selection}) in zero-shot prompting (see Section \ref{sec:zerovsfewshot}). Passes 
alternate between SR and WR improvement steps. The optimal order and number of passes are empirically determined by ablation studies (Section \ref{results}).
\begin{enumerate}
    \item \textbf{SR Improvement}: Maps generic speaker labels (SPEAKER\_00, SPEAKER\_01) to clinical roles (Patient, Neurosurgeon, etc.) based on conversational patterns and medical terminology.
    \item \textbf{WR Improvement}: Corrects ASR errors using the identified clinical context, and pseudonymizes personal names while preserving spontaneous speech markers.
    \item \textbf{SR Refinement}: Re-evaluates speaker attributions using the corrected transcript to resolve ambiguities that were obscured by initial ASR errors.
    \item \textbf{WR Refinement}: Re-evaluates lexical corrections using the refined speaker attributions to resolve remaining transcription ambiguities that were obscured by initial speaker errors.
\end{enumerate}
This architecture is designed based on three key principles validated through systematic ablation studies presented in Section~\ref{results}. This architecture can be interpreted as a form of alternating contextual refinement, where speaker role inference and lexical correction act as mutually constraining objectives. Rather than treating speaker attribution and lexical correction as independent tasks, the model iteratively stabilizes each dimension using the contextual signal provided by the other. All experiments use Qwen3-Next-80B, a recent large-scale open-weight model optimized for high-reasoning tasks \cite{qwen3technicalreport}.

\subsection{Prompts}

The performance of the LLM post-processing stage relies on specialized system prompts designed to provide the models with both clinical domain expertise and structural output constraints. Prompts are adapted for each medical domain (AN vs SP) and for each processing pass type (SR vs WR).

\subsubsection{Speaker Recognition (SR) Pass}

The SR prompt uses role prompting to assign the model a domain-specific expert role (e.g., ``Expert in neurosurgical consultation analysis'' for the AN dataset, and ``Expert in analyzing telephone conversations for the suicide prevention hotline'' for the SP dataset) and describes the expected social and clinical dynamics of the conversation. The model is instructed to:

\begin{itemize}
    \item Map generic labels (SPEAKER\_00, SPEAKER\_01) to clinical roles based on technical vocabulary, question-answer patterns, and conversational hierarchy, explicitly correcting any inconsistencies when spoken content suggests a mismatch with previously assigned roles.
    
    For example, in the AN dataset: \textit{``The neuropsychologist: gives instructions, asks test 
    questions (naming images, completing sentences, etc.) / The patient: responds to tests, may express fatigue or discomfort"}.
    \item Maintain longitudinal consistency: each unique SPEAKER\_XX label should ideally map to a single role throughout the recording, following the instruction: \textit{``each SPEAKER\_XX keeps the 
    SAME role from beginning to end"}.
\end{itemize}

For the SP dataset, the prompt includes an additional verbatim reference to identify the automated voice system that typically opens calls, ensuring that it is not confused with human speakers.

\subsubsection{Word Recognition (WR) Pass}

The WR pass focuses on refining the ASR output while preserving the spontaneous nature of speech. The prompt instructs the model to:

\begin{itemize}
    \item Correct systematic ASR errors using a domain-specific ``correction dictionary'' derived from preliminary analysis of WhisperX's most frequent errors on French medical speech \cite{bertin2025transcription}, and resolve phonetic hallucinations using the clinical context established by preceding SR passes. 
    
    For example: \textit{``je tue toi" $\rightarrow$ ``je tutoie"}, \textit{``crise épistique" $\rightarrow$ ``crise d'épilepsie"}, \textit{``c'est son précédent" $\rightarrow$ ``ses antécédents"}.
    \item Pseudonymize personal identifiable information by replacing all proper names with a generic token, following the instruction: \textit{``always replace ALL proper names with `name'"}.
    \item Preserve spontaneous speech markers: hesitation markers (``euh'', ``hm'') and informal expressions (``ouais'', ``bah'') must be maintained to avoid over-summarization, a common tendency in LLMs, following the instruction: \textit{``preserve oral speech markers (`euh', `hm', `hein', `ouais', `bah')"}.
    
    This is particularly important for mental health applications, where prosodic markers and hesitations convey emotional states essential for clinical assessment \cite{Cummins2015ASpeech,marie2025acoustic}.
\end{itemize}

Both prompt types include explicit formatting instructions requiring the model to output structured segment lists. Each segment must preserve its original format:
\begin{verbatim}
[start_time] - [end_time] [Speaker_Label]
[transcribed text]
\end{verbatim}
This prevents LLM failure modes such as generating free-form summaries, merging adjacent segments, or producing narrative-style text that cannot be automatically parsed and aligned with the original audio timestamps.

\subsection{Hyperparameter Optimization}

To justify each design decision, the following alternative configurations are evaluated.

\subsubsection{Model Selection}
\label{sec:model_selection}
To identify the optimal LLM for our pipeline, three models representing different accessibility and scale trade-offs are compared in a single-pass post-processing configuration where the model executes joint SR and WR in a single inference:
\begin{itemize}
    \item \textbf{GPT-4o-mini} (subsequently GPT4omini): A state-of-the-art commercial model from OpenAI, chosen for its high performance-to-latency ratio and proven robustness in instruction-following tasks. 
    \item \textbf{Qwen3-Next-80B-A3B-Instruct-AWQ-4bit} (subsequently Qwen80B): A large-scale open-weight model designed for high-reasoning tasks. A 4-bit AWQ quantized version is used to evaluate whether massive open-source models can match or exceed commercial APIs in specialized medical contexts while enabling local deployment.
    \item \textbf{Qwen3-VL-8B-Instruct-FP8} (subsequently QwenVL): Although primarily a Vision-Language model, this 8B parameter version is included to assess whether compact models, potentially easier to deploy in resource-constrained clinical settings, can provide sufficient correction capabilities when restricted to text-only input.
\end{itemize}
The single-pass configuration provides a controlled comparison without any architectural complexity, allowing for direct assessment of each model's SR and WR capabilities. All LLM inferences were conducted with temperature set to 0, 
ensuring fully deterministic outputs. Transcripts contain only verbal exchanges, without medical records or administrative identifiers. Therefore, experiments did not involve the transmission of identifiable data to external servers. Based on this evaluation (Section~\ref{model_selection}), the optimal model is then used for all subsequent experiments.

\subsubsection{Zero vs Few-shot Prompting}
\label{sec:zerovsfewshot}
To evaluate the impact of in-context examples on post-processing stability and accuracy, two prompting strategies are compared using Qwen80B. The comparison uses a sequential two-pass post-processing : first, the LLM maps generic speaker labels to clinical roles (SR Pass), then corrects lexical errors using the identified roles as context (WR Pass).
\begin{itemize}
    \item \textbf{Zero-shot}: The model receives only task instructions without examples, relying on its pre-trained knowledge of medical terminology and conversational structures.
    \item \textbf{Few-shot}: The prompt is augmented with a representative example of the target dataset (AN or SP), illustrating expected medical vocabulary, speaker role mappings, and segment formatting.
\end{itemize}

\subsubsection{SR vs WR-led Prompting}

To determine the optimal ordering of the SR and WR passes, two sequential strategies are compared using Qwen80B in zero-shot mode in Two-pass and Three-pass configurations.
\begin{itemize}
    \item \textbf{SR-led}: The first pass maps generic speaker labels (SPEAKER\_00, SPEAKER\_01) to clinical roles (Patient, Neurosurgeon, etc.), providing contextual information for subsequent lexical correction. In the Three-pass architecture, this strategy alternates: Pass 1 (SR Improvement) $\rightarrow$ Pass 2 (WR Improvement) $\rightarrow$ Pass 3 (SR Refinement).
    \item \textbf{WR-led}: The first pass resolves ASR errors and standardizes terminology before speaker attribution. In the Three-pass architecture, this alternates between: Pass 1 (WR Improvement) $\rightarrow$ Pass 2 (SR Improvement) $\rightarrow$ Pass 3 (WR Refinement).
\end{itemize}
This comparison evaluates whether it is more effective to first identify speaker roles and then use this context to guide lexical correction (SR-led) or to first clean the transcript and then use the corrected text to improve speaker attribution (WR-led).
Additionally, a joint multi-pass baseline is evaluated, in which a single prompt performing SR and WR simultaneously is repeated N=2 and N=3 times without task separation, to verify that improvements stem from task separation rather than repeated processing.  

\subsubsection{Iteration Depth}

To identify the optimal number of SR-WR cycles before diminishing returns or instability, configurations ranging from one to seven passes are evaluated using Qwen80B in zero-shot SR-first mode. 

The Single-pass configuration performs joint WR and SR in a single inference, testing whether Qwen80B can improve upon the baseline without iterative refinement. 
The Two-pass approach splits the task into sequential stages: an initial SR Pass maps speaker labels to clinical roles, followed by a WR Pass that leverages this role information to guide lexical modifications. 
The Three-pass architecture adds a SR refinement stage, re-evaluating speaker attributions using the now-corrected transcript to resolve ambiguities that were obscured by initial ASR errors. 
Configurations with four to nine passes extend this alternating pattern, adding successive WR and SR stages to test whether further iteration continues to improve performance or introduces instability, until computation time exceeds real audio time with Real-Time-Factor$>$1 (see Section \ref{sec:rtf}).

\subsection{Evaluation Metrics}
\subsubsection{Word Diarization Error Rate (WDER)}

Given the dual objectives of accurate lexical transcription and reliable speaker attribution, Word Diarization Error Rate (WDER) is selected as the primary evaluation metric. WDER is a word-level speaker attribution metric that measures the proportion of words with incorrect speaker tags among aligned words. In our implementation \cite{Shafey2019Joint,Tran2022Automatic}, lexical and speaker errors are combined as:

\begin{equation}
\text{WDER} = \text{WER} + \frac{W_{\text{correct but wrong speaker}}}{N}
\end{equation}
where
\begin{itemize}
    \item $W_{\text{correct but wrong speaker}}$ = number of correctly transcribed words assigned to the wrong speaker,
    \item $N$ = total number of words in the reference transcript.
\end{itemize}

Unlike pure diarization metrics (DER) that penalize temporal boundary shifts regardless of semantic correctness, or pure transcription metrics (WER) that ignore speaker attribution, WDER captures both dimensions simultaneously. Moreover, LLMs can improve semantic speaker attribution while slightly altering temporal boundaries. WDER rewards correct word-level assignments even when segment boundaries are redistributed based on conversational logic.
Pure speaker-level WDER (e.g. proportion of misattributed words among aligned words) is also used in the literature \cite{Paturi2023Lexical,Paturi2024AG-LSEC} but this version allows direct comparison with WER.

Diarization Error Rate (DER) is reported alongside WDER and WER for completeness, but is not used as a primary evaluation criterion. Reference transcriptions contain no manually annotated temporal boundaries : segment timestamps were estimated by WhisperX, creating an evaluation where both the baseline and the reference share the same alignment bias. DER therefore reflects alignment artifacts rather than true diarization quality. Similarly, concatenated minimum-permutation WER (cpWER) \cite{von2025word}, the community standard for the evaluation of speaker-attributed ASR in meeting transcription benchmarks \cite{carletta2007unleashing,watanabe2020chime}, is designed for systems producing full speaker-separated streams with temporal boundaries. WDER at word level avoids the temporal confound via lexical alignment with the linguist-produced reference, making it the appropriate primary metric for our post-processing evaluation context.

\subsubsection{Format Errors}
Files that cannot be parsed as valid output due to format violations are defined as ``outliers'' (also referred to as parsing failures).

\subsubsection{Real-Time Factor (RTF)}
\label{sec:rtf}
The ratio of processing time to audio duration is called Real-Time Factor (RTF), and is used specifically in the iteration depth analysis (Section~\ref{iteration_depth}) to quantify the computational cost of multi-pass strategies.
RTF is defined as:

\begin{equation}
\text{RTF} = \frac{T_{\text{processing}}}{T_{\text{audio}}}
\end{equation}

where $T_{\text{processing}}$ is the total wall-clock time required to process a recording and $T_{\text{audio}}$ is the duration of the corresponding audio file. An RTF below 1 indicates faster-than-real-time processing.

\subsubsection{Statistical significance}
Statistical significance is assessed using the Wilcoxon signed-rank test \cite{woolson2007wilcoxon}, a non-parametric test for paired comparisons. For each pairwise comparison (e.g., 3P-S vs Baseline), the WDER difference for each individual recording is computed. The test then evaluates whether the median of these paired differences is significantly different from zero, indicating that one configuration consistently outperforms the other across recordings. Comparisons are conducted separately for each dataset (AN and SP) due to their distinct conversational characteristics. Results are considered statistically significant when $p \leq 0.05$.

\section{Results}
\label{results}

\subsection{Model selection}
\label{model_selection}
To identify the optimal model for our application, three LLM candidates are first compared in a single-pass configuration where the model performs joint SR and WR in a single inference. Table~\ref{tab:llm} presents the comparative performance on both clinical datasets.

\begin{table}[ht!]
  \caption{Model comparison in single-pass configuration.}
  \label{tab:llm}
  \centering
  \begin{tabular}{lcccccccc}
    \toprule
    & \multicolumn{3}{c}{\textbf{AN}} & \multicolumn{3}{c}{\textbf{SP}} & & \\
    \cmidrule(lr){2-4} \cmidrule(lr){5-7}
    \textbf{Model} & \textbf{WER} & \textbf{DER} & \textbf{WDER} & \textbf{WER} & \textbf{DER} & \textbf{WDER} & \textbf{Avg} & \textbf{Outliers}\\
    \midrule
    Baseline  & 32.67 & 24.96 & 43.03 & 30.03 & 28.28 & 42.45 & 42.74 & 0 \\
    GPT4omini & 32.62 & 24.96 & 42.93 & 30.04 & 28.21 & 42.44 & 42.69 & 0 \\
    QwenVL    & 34.68 & 26.02 & 45.46 & 30.43 & 33.17 & 42.55 & 44.01 & 3 \\
    Qwen80B   & 32.66 & 24.96 & 43.02 & 30.03 & 28.19 & 42.44 & 42.73 & 0 \\
    \bottomrule
  \end{tabular}
  \\Avg: mean WDER across both datasets.
\end{table}

\subsection{Zero vs Few-shot prompting}
The effect of augmenting prompts with in-context examples on performance and stability is evaluated with Qwen80B. Table~\ref{tab:zsfs} compares zero-shot and few-shot prompting strategies in a two-stage workflow (SR $\rightarrow$ WR).

\begin{table}[ht!]
  \caption{Zero-shot vs Few-shot prompting with Qwen80B.}
  \label{tab:zsfs}
  \centering
  \begin{tabular}{lcccccccc}
    \toprule
    & \multicolumn{3}{c}{\textbf{AN}} & \multicolumn{3}{c}{\textbf{SP}} & & \\
    \cmidrule(lr){2-4} \cmidrule(lr){5-7}
    \textbf{Strategy} & \textbf{WER} & \textbf{DER} & \textbf{WDER} & \textbf{WER} & \textbf{DER} & \textbf{WDER} & \textbf{Avg} & \textbf{Outliers}\\
    \midrule
    Baseline  & 32.67 & 24.96 & 43.03 & 30.03 & 28.28 & 42.45 & 42.74 & 0 \\
    Zero-shot & 32.68 & 27.29 & 43.76 & 29.99 & 26.89 & 39.96 & 41.86 & 0 \\
    Few-shot  & 32.40 & 25.67 & 42.12 & 30.03 & 26.32 & 40.44 & 41.28 & 1 \\
    \bottomrule
  \end{tabular}
\end{table}

\subsection{SR vs WR-led prompting}
The impact of pass ordering and task separation is evaluated to assess if SR should precede WR (as in our proposed method), or if lexical cleaning should occur first. Table~\ref{tab:dvsc} compares both orderings across two-pass and three-pass configurations. All configurations produced zero outliers.

\begin{table}[ht!]
  \caption{SR-led vs WR-led pass ordering with Qwen80B, including 
  joint multi-pass baseline.}
  \label{tab:dvsc}
  \centering
  \begin{tabular}{lccccccccc}
    \toprule
    & & \multicolumn{3}{c}{\textbf{AN}} & \multicolumn{3}{c}{\textbf{SP}} & & \\
    \cmidrule(lr){3-5} \cmidrule(lr){6-8}
    \textbf{Strategy} & \textbf{N} & \textbf{WER} & \textbf{DER} & \textbf{WDER} & \textbf{WER} & \textbf{DER} & \textbf{WDER} & \textbf{Avg} & \textbf{Outliers}\\
    \midrule
    Baseline   & 0 & 32.67 & 24.96 & 43.03 & 30.03 & 28.28 & 42.45 & 42.74 & 0 \\
    Joint-pass & 2 & 32.68 & 24.96 & 43.04 & 30.03 & 28.19 & 42.43 & 42.74 & 0 \\
    Joint-pass & 3 & 32.69 & 24.96 & 43.05 & 30.03 & 28.19 & 42.43 & 42.74 & 0 \\
    SR-led     & 2 & 32.68 & 27.29 & 43.76 & 29.99 & 26.89 & 39.96 & 41.86 & 0 \\
    WR-led     & 2 & 32.71 & 27.84 & 44.62 & 29.99 & 26.55 & 39.72 & 42.17 & 0 \\
    SR-led     & 3 & 32.68 & 27.45 & 43.79 & 29.99 & 26.32 & 39.57 & 41.68 & 0 \\
    WR-led     & 3 & 32.72 & 27.52 & 43.96 & 30.03 & 27.09 & 41.04 & 42.50 & 0 \\
    \bottomrule
  \end{tabular}
\end{table}

\subsection{Iteration depth}
\label{iteration_depth}
The optimal number of iterative refinement cycles is determined with Table~\ref{tab:wder_rtf}, which evaluates configurations ranging from one to seven passes, measuring WDER improvement, output stability (outliers) and computational cost (Real-Time Factor).

\begin{table}[ht!]
  \caption{Iteration depth analysis with Qwen80B.}
  \label{tab:wder_rtf}
  \centering
  \begin{tabular}{lcccccccc}
    \toprule
    & \multicolumn{3}{c}{\textbf{AN}} & \multicolumn{3}{c}{\textbf{SP}} & & \\
    \cmidrule(lr){2-4} \cmidrule(lr){5-7}
    \textbf{N pass} & \textbf{WER} & \textbf{DER} & \textbf{WDER} & \textbf{WER} & \textbf{DER} & \textbf{WDER} & \textbf{Avg} & \textbf{RTF}\\
    \midrule
    0 & 32.67 & 24.96 & 43.03 & 30.03 & 28.28 & 42.45 & 42.74 & 0.09 \\
    1 & 32.66 & 24.96 & 43.02 & 30.03 & 28.19 & 42.44 & 42.73 & 0.12 \\
    2 & 32.68 & 27.29 & 43.76 & 29.99 & 26.89 & 39.96 & 41.86 & 0.22 \\
    3 & 32.68 & 27.45 & 43.79 & 29.99 & 26.32 & 39.57 & 41.68 & 0.32 \\
    4 & 32.68 & 27.48 & 43.96 & 29.99 & 26.46 & 39.33 & 41.65 & 0.45 \\
    5 & 32.40 & 27.90 & 44.38 & 29.99 & 28.15 & 40.75 & 42.57 & 0.57 \\
    6 & 32.81 & 27.32 & 43.32 & 29.99 & 26.44 & 39.76 & 41.54 & 0.68 \\
    7 & 33.01 & 27.63 & 42.65 & 29.99 & 26.41 & 39.89 & 41.27 & 0.78 \\
    8 & 32.70 & 27.27 & 43.94 & 29.99 & 26.15 & 39.73 & 41.84 & 0.89 \\
    9 & 32.68 & 27.16 & 44.12 & 29.99 & 27.91 & 40.81 & 42.47 & 1.04 \\
    \bottomrule
  \end{tabular}
  \\RTF: Real-Time Factor. N pass=0 corresponds to the Baseline.
\end{table}

\subsection{Statistical significance}
To quantify the robustness of our chosen method for our datasets, pairwise statistical comparisons are conducted between the Three-pass SR-led method (3P-S) and all alternative configurations evaluated in the preceding ablation studies. Table~\ref{tab:stat} presents Wilcoxon signed-rank test results comparing 3P-S to baseline, single-pass models, alternative prompting strategies, alternative pass orderings, and alternative iteration depths. Negative $\Delta$WDER values indicate improvements by the proposed method, and significant comparisons (p $<$ 0.05) along with improvements are highlighted in bold.

\begin{table}[ht!]
  \caption{Statistical comparison of 3P-S vs alternatives.}
  \label{tab:stat}
  \centering
  \small
  \begin{tabular}{l@{\hspace{0.8em}}cc@{\hspace{1.5em}}cc}
    \toprule
    & \multicolumn{2}{c}{\textbf{AN}} & \multicolumn{2}{c}{\textbf{SP}} \\
    \cmidrule(lr){2-3} \cmidrule(lr){4-5}
    \textbf{vs 3P-S} & $\boldsymbol{\Delta}$\textbf{WDER} & \textbf{p-value} & $\boldsymbol{\Delta}$\textbf{WDER} & \textbf{p-value} \\
    \midrule
    Baseline         & -0.76 & 1.00 & \textbf{+2.88*} & \textbf{0.02} \\
    1P-S (GPT4omini) & -0.86 & 0.77 & \textbf{+2.87*} & \textbf{0.02} \\
    1P-S (QwenVL)    & +0.98 & 0.13 & \textbf{+4.02*} & \textbf{0.02} \\
    1P-S             & -0.77 & 1.00 & \textbf{+2.87*} & \textbf{0.02} \\
    2P-S             & -0.03 & 0.31 & +0.39  & 0.40 \\
    2P-S-FS          & -1.67 & 0.16 & +0.87  & 0.82 \\
    3P-W             & +0.17 & 1.00 & +1.46  & 0.44 \\
    4P-S             & +0.17 & 0.69 & -0.24  & 0.11 \\
    5P-S             & -0.07 & 0.20 & +1.18  & 0.51 \\
    6P-S             & -0.47 & 0.22 & +0.18  & 0.83 \\
    7P-S             & -1.14 & 0.22 & +0.32  & 0.61 \\
    8P-S             & +0.15 & 0.94 & +0.16  & 0.88 \\
    9P-S             & +0.33 & 0.84 & +1.24  & 0.28 \\
    \bottomrule
  \end{tabular}
  \\nP-X = n passes with strategy X (S=SR-led, W=WR-led, FS=Few-Shot).
  \\Models in parentheses indicate single-pass with that specific model. No parentheses = Qwen80B.
  \\$\Delta$WDER (pp) = mean of per-recording paired absolute differences (WDER\textsubscript{alternative} $-$ WDER\textsubscript{3P-S}). Positive values indicate improvement by 3P-S.
  \\ * significant (p $\leq$ 0.05) improvements.
\end{table}

\subsection{WDER Decomposition}

\begin{figure}[ht!]
\centering
\caption{Decomposition of WDER}
\includegraphics[width=0.8\linewidth]{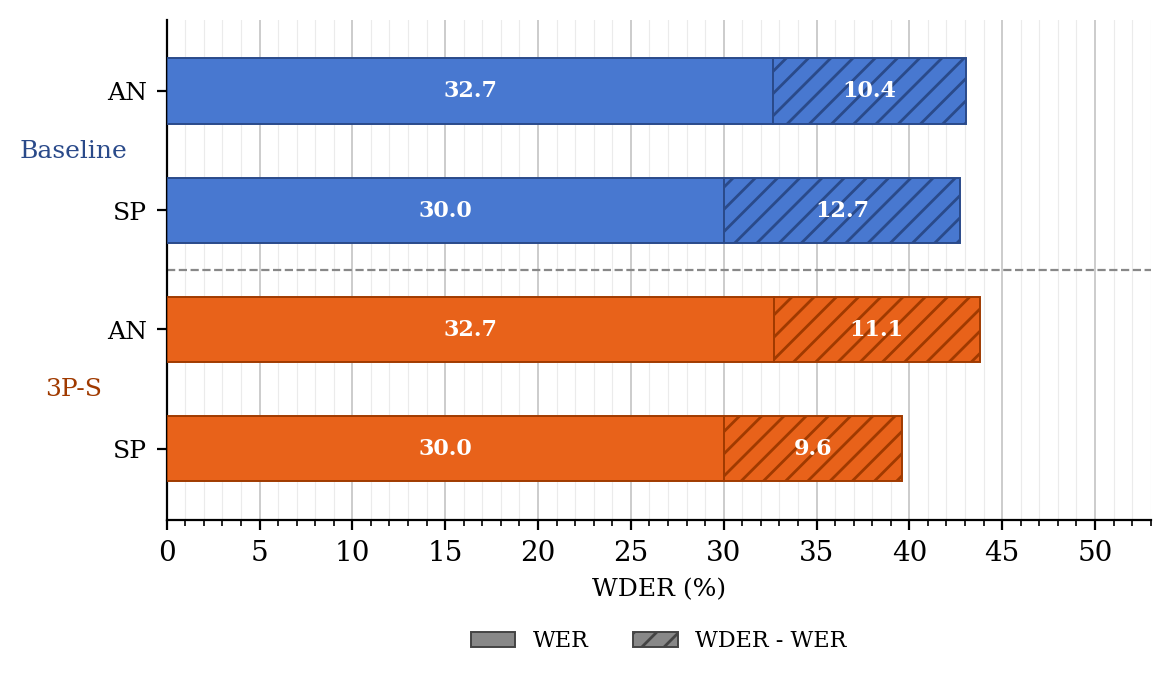}

\label{fig:decomposition_WDER}
\end{figure}

Figure \ref{fig:decomposition_WDER} presents the decomposition of WDER into lexical error rate (WER) and speaker attribution error (WDER-WER) for the Baseline and 3P-S configurations on both datasets.

\subsection{Qualitative analysis}
\begin{figure}[ht!]
\centering
\caption{Representative examples of qualitative improvements.}
\includegraphics[width=0.6\linewidth]{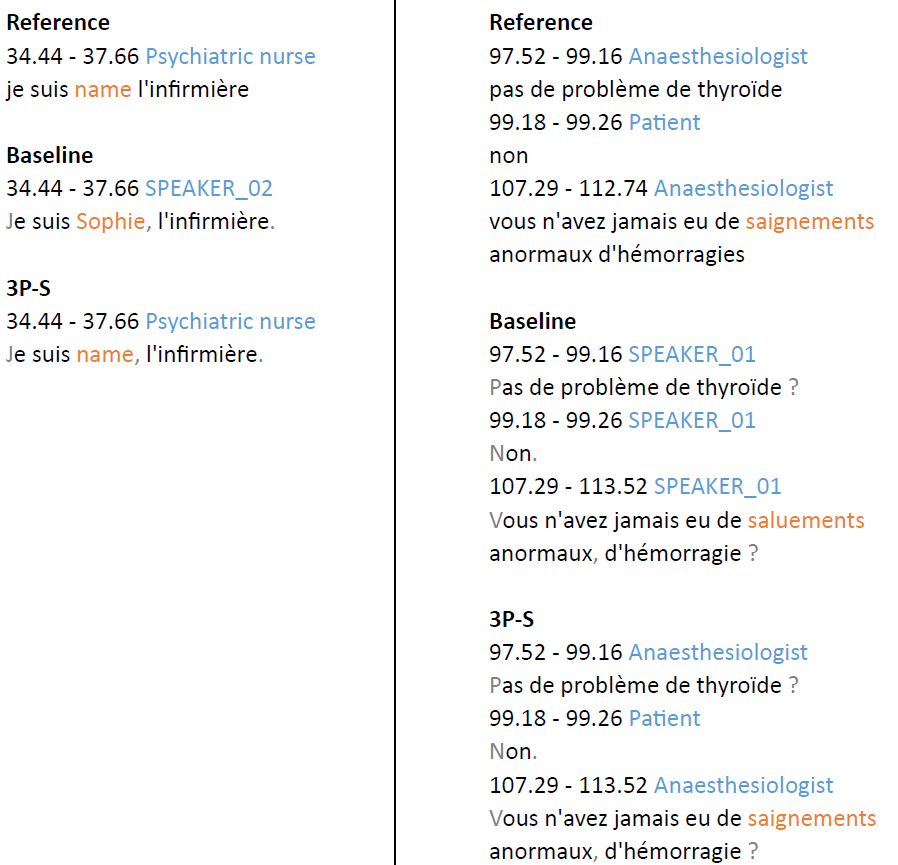}

\label{fig:qualitative_examples}
\end{figure}

Beyond quantitative metrics, systematic improvements in clinical usability were observed in both datasets. The proposed 3P-S method presents semantic speaker labels replacing generic tags and preservation of pseudonymization tokens. Representative examples illustrate these improvements in Figure \ref{fig:qualitative_examples}. Baseline outputs use generic SPEAKER\_XX tags, while 3P-S provides role-specific labels and distinguishes neurosurgeon from family member (\textit{Proche}), a critical distinction absent in Baseline's generic labels. 3P-S preserves pseudonymization tokens (\texttt{name}) that Baseline replaces with real heard names. 

Preservation of spontaneous speech markers was verified across all recordings: a systematic count of nine marker types (\textit{euh}, \textit{hm}, \textit{hein}, \textit{ouais}, \textit{bah}, \textit{ah}, \textit{oh}, \textit{ben}, \textit{bon}) confirmed zero suppressions in both datasets.

\section{Discussion}

\subsection{Model selection}
Table~\ref{tab:llm} presents the comparative performance of three LLM candidates in both clinical datasets. QwenVL shows systematic degradation relative to baseline on AN (WDER +2.43 points) and produces three unparseable outputs. This was expected, as QwenVL is optimized for vision-language tasks and the 8B parameter scale is insufficient for complex medical dialogue reasoning \cite{Bai2023Qwen-VL:,Singhal2022Large}. This result demonstrates that compact models cannot replace larger instruction-tuned LLMs for medical transcription post-processing, suggesting that model scale is a critical factor for this task.

GPT4omini achieves improvements on both datasets (avg relative WDER reduction of 0.1\%) with zero outliers, demonstrating robust baseline maintenance. However, its commercial nature limits deployment flexibility and incurs per-inference costs.

Qwen80B matches baseline performance within measurement precision (avg WDER reduction of 0.02\% relative) while offering three decisive advantages: (1) open-weights licensing enabling local deployment without per-inference costs, (2) stability with zero outliers across both datasets, and (3) cross-domain consistency comparable to GPT4omini (0.58 WDER difference between datasets vs 0.49 for GPT4omini).

Based on these results, Qwen80B is selected for all subsequent experiments. While single-pass improvements are minimal, this establishes a stable foundation for iterative refinement strategies explored in subsequent sections. The comparable performance between GPT4omini and Qwen80B suggests that open-source models have reached sufficient capability for medical transcription post-processing tasks, provided that sufficient clinical context is given with structured prompting and iterative refinement.

\subsection{Zero vs Few-shot prompting}

Table~\ref{tab:zsfs} compares zero-shot and few-shot prompting strategies for Qwen80B in a Two-pass SR-led configuration. Few-shot prompting achieves lower average WDER (41.28 vs 41.86 for zero-shot, -1.4\% relative improvement), with particularly strong gains in AN (WDER 42.12 vs 43.76, -3.7\% relative).

However, this marginal improvement comes at the cost of one unparseable output (outlier) in the few-shot condition, representing a production stability risk. Although few-shot prompting can improve domain adaptation, it also risks introducing output format instability or over-fitting to specific conversational patterns \cite{Zhao2021Calibrate}. 

Given that zero-shot already delivers substantial improvements over baseline (avg WDER 41.86 vs 42.74, -2.1\% relative) with perfect output parseability, the additional complexity and instability introduced by few-shot examples is not justified.
Zero-shot prompting is therefore retained for the proposed method, prioritizing robustness for clinical deployment over marginal WDER gains.

\subsection{SR vs WR-led prompting}

Table~\ref{tab:dvsc} evaluates the impact of pass ordering across both Two-pass and Three-pass architectures. In the Two-pass setting, WR-led achieves marginally lower WDER on SP (39.72 vs 39.96), but this advantage reverses in the Three-pass setting where SR-led outperforms (39.57 vs 41.04, -3.6\% relative).

Repeating the joint prompt N=2 and N=3 times does not improve the single-pass baseline. Average WDER remains equivalent across all joint configurations, and WER and DER do not show reduction. This confirms that the alternating SR/WR architecture improves performance through task separation, not through repeated processing. When speaker attribution and lexical correction are addressed in a single prompt, the model fails to improve either task.

More important, SR-led ordering demonstrates superior cross-dataset stability. The performance gap between datasets is 3.80 WDER points for 2P SR-led versus 4.90 for 2P WR-led. This pattern intensifies in Three-pass, where SR-led maintains 4.22 point gap versus 2.92 for WR-led, but the absolute performance on SP is substantially better (39.57 vs 41.04).

The advantage of SR-led likely comes from LLMs' limited impact on pure lexical correction. Models pre-trained on written text excel at speaker role attribution when provided with conversational structure, but struggle to outperform domain-tuned ASR on phonetic disambiguation without first understanding who is speaking.

SR-led ordering is therefore preferred, as it provides the best compromise between per-dataset performance and cross-domain generalization, achieving consistent improvements in SP while maintaining stability in AN.

\subsection{Iteration depth}

Table~\ref{tab:wder_rtf} reveals a non-monotonic relationship between iteration depth and performance. Single-pass processing maintains Baseline quality (avg WDER 42.73 vs 42.74) with minimal computational overhead (RTF 0.12 vs 0.09), confirming that Qwen80B's instruction-following capacity is sufficient for joint WR-SR but does not improve upon the pipeline without iterative refinement.

Two-pass and Three-pass configurations deliver consistent improvements on SP (WDER 39.96 and 39.57 respectively, vs baseline 42.45), with Three-pass achieving the best WDER on this dataset. The marginal gain from Four-pass (39.33) is offset by a 40\% longer processing time (RTF 0.45 vs 0.32) for only 0.24 WDER reduction. This cost-benefit ratio worsens further at higher depths.

Beyond four passes, stability degrades: Five-pass produces one outlier and shows WDER regression on both datasets relative to Three-pass. Six-pass and Seven-pass show high variance, with Seven-pass achieving competitive average WDER (41.27) but at 2.4× baseline computational cost (RTF 0.78 vs 0.32 for Three-pass).

Three-pass represents the optimal balance for our datasets. It achieves 97.6\% of Four-pass WDER performance (41.68 vs 41.65) while maintaining perfect output stability with zero outliers, and 29\% lower computational cost (RTF 0.32 vs 0.45). Although Four-pass shows marginally better WDER on SP (39.33 vs 39.57), this 0.24-point difference is not statistically significant (p=0.11, Table \ref{tab:stat}) and does not justify the increased computational overhead for offline batch processing scenarios. For applications requiring maximum accuracy regardless of cost, Four-pass may be preferred.

These observations are further supported by statistical comparisons presented in the following section.

\subsection{Statistical significance}
Table~\ref{tab:stat} presents pairwise statistical comparisons of the proposed Three-pass SR-led method (3P-S) compared to alternative configurations. Given the exploratory nature of these comparisons across multiple configurations, uncorrected p-values are reported to avoid inflating Type II error. Borderline significant results (0.01 $<$ p $<$ 0.05) should be interpreted with appropriate caution. 
On SP, 3P-S achieves statistically significant improvements (p $<$ 0.05 with Wilcoxon signed-rank test) over baseline (mean $\Delta$WDER = -7.30 points) and all Single-pass configurations regardless of model choice (GPT4omini: -7.20, QwenVL: -10.40, Qwen80B: -7.30).

SP conversations involve higher lexical diversity (type-token ratio 0.49 vs 0.30) than the more structured AN consultations. This likely makes SP more sensitive to SR refinement, as LLMs can leverage contextual emotional cues and conversational dynamics to disambiguate speaker attribution. In contrast, AN's more formal consultation structure provides clearer speaker boundaries even in the baseline, leaving less room for improvement through iterative refinement.

Figure \ref{fig:decomposition_WDER} confirms this interpretation : WER remains unchanged between Baseline and 3P-S on both datasets (AN: 32.7\%, SP: 30.0\%), indicating that all WDER differences are driven by changes in speaker attribution. On SP, the attribution component decreases from 12.7 to 9.6 percentage points, which is responsible for the observed WDER reduction. On AN, the marginal increase in the attribution component indicates that the LLM introduces minor speaker assignment errors on a dataset where turn-taking is already well-structured.

Moreover, no configuration shows statistically significant degradation relative to 3P-S on either dataset, confirming that the proposed method does not introduce regressions. In AN, 3P-S maintains performance within statistical equivalence of Baseline ($\Delta$WDER = +1.70, p = 1.00), indicating dataset-specific effectiveness rather than universal improvement.

Comparisons between 3P-S and other multi-pass depths (2P, 4P, 5P, 6P, 7P) show no significant differences (all p $>$ 0.10), suggesting that beyond the initial two-pass threshold, performance gains stabilize within the statistical noise of our sample size. The lack of significance for 4P-S vs 3P-S ($\Delta$WDER = +0.60 on SP, p = 0.11) reinforces that the marginal 0.60-point improvement observed in Table~\ref{tab:wder_rtf} does not justify the 40\% increase in computational cost (RTF 0.45 vs 0.32).

The statistically significant improvements on SP, combined with stability on AN and zero production failures (outliers), support the adoption of 3P-S for clinical deployment across both our conversational contexts, with larger gains expected in emotionally charged interactions with higher lexical diversity.
However, absolute WDER values remain high (around 40\%), preventing fully autonomous clinical deployment. The objective of this feasibility study is not to replace human correction, but to target the most time-consuming aspect of manual editing for clinical linguists by improving speaker attribution.

\section{Perspectives}

Several research directions could extend this work. 
Recent work suggests that Chain-of-Thought (CoT) prompting, where the LLM is instructed to explicitly verbalize its reasoning process before making corrections, can improve transcription accuracy on English medical consultations \cite{adedeji2024sound}. Integrating CoT reasoning into the SR pass could help the model justify speaker role assignments based on conversational cues, potentially reducing attribution errors in ambiguous cases. 

The combination of outputs from multiple LLMs by majority vote or confidence-weighted averaging can reduce individual model hallucinations and improve robustness \cite{Efstathiadis2024LLM-based}. Applying ensemble techniques to the proposed pipeline could reduce outlier risk and improve performance on both datasets simultaneously.

A key limitation of this study is the modest sample size (10 AN and 18 SP conversations), which limits statistical power and generalizability claims. Future work should validate the proposed architecture on larger corpora across additional French medical applications (e.g., emergency medicine, general practice) to assess robustness beyond the two domains studied here. Multi-site validation would also help establish whether the observed dataset-specific effects (improvements on SP but not AN) reflect fundamental differences in conversation structure or artifacts of our specific recording conditions.

The correction dictionary used in WR passes was derived from the study datasets themselves, representing a form of in-domain supervision that may overestimate generalizability. Future work should evaluate whether a domain-agnostic dictionary, or automatic dictionary induction from held-out data, yields comparable improvements.

The chunking strategy (500 segments per chunk) was not systematically optimized. The effect of smaller context windows on LLM attention and correction quality remains unexplored and represents an additional hyperparameter for future investigation.

The baseline diarization component (Pyannote 3.1) could be replaced by more recent systems such as DiariZen \cite{polok2025but}, whose stable benchmarks were released recently. However, all published evaluations of DiariZen on French data are strictly limited to two-speaker sessions, whereas our datasets involve up to four simultaneous speakers (automated voice, nurse(s), patient in SP; neurosurgeon, neuropsychologist, patient, family member in AN). Performance on French multi-speaker clinical speech therefore remains undocumented, and DiariZen represents a natural direction for future work once its multi-speaker French capabilities are established.
Comparison with domain-adapted ASR models fine-tuned on French medical speech would further contextualize the contribution of LLM post-processing.

Future work should assess whether improved speaker attribution reduces manual correction effort for clinical linguists. 
A question that remains open is also whether better transcriptions improve automatic classification performance in tasks related to datasets, such as suicide risk assessment or surgical outcome prediction. Moreover, tighter ASR-LLM integration, such as feeding n-best hypotheses to enable contextual re-ranking, could further improve correction quality, with clinical utility as the ultimate validation criterion.

\section{Conclusion}

This work proposes a N-pass LLM post-processing architecture for French clinical interview transcription, validated through systematic ablation studies on two distinct clinical datasets: suicide prevention telephone counseling and preoperative awake neurosurgery consultations.

Large-scale open-source models (Qwen3-Next-80B, 80B parameters) match commercial API performance while enabling local deployment, addressing privacy and cost constraints critical for clinical applications. This approach proposes a reproducible framework for other under-resourced languages in medical contexts, reducing dependence on English-centric commercial platforms.
Moreover, zero-shot prompting ensures production stability (zero output failures), SR-led ordering provides superior cross-dataset generalization, and three iterative passes represent the optimal balance between accuracy gains and computational cost. 
The proposed method achieves statistically significant improvements in word-level speaker attribution in emotionally charged conversations (relative WDER reduction of 6.8\%, p $<$ 0.05 on SP) while maintaining stability in structured medical consultations.

These results are achieved with acceptable computational cost (RTF 0.32, approximately a third of real-time) and high output stability, suggesting feasibility for offline clinical deployment in conversation analysis, clinical documentation, and research applications requiring speaker-attributed transcriptions. Clinical validation with end-users and integration into existing workflows remain necessary steps before operational deployment.

Future work should explore advanced prompting strategies (chain-of-thought reasoning), LLM ensemble methods, and tighter ASR-LLM integration to further improve robustness and extend applicability to real-time scenarios and various clinical contexts.

% \section{Acknowledgments}

% This research was approved by an Ethic Committee and the institutional review boards of participating hospitals. All participants provided informed consent for their conversations to be used for research purposes. Audio recordings were pseudonymized prior to processing, and all personally identifiable information was replaced with generic tokens in the final transcripts. 

% This work was funded by the Brittany Region (France) through the doctoral program ``Allocations de Recherche Doctorale'' (ARED). 

% Ambre Marie designed the experiments, conducted the analyzes, and drafted the manuscript. Thomas Bertin created the reference transcriptions. Gwenolé Quellec, Guillaume Dardenne and Thomas Bertin supervised the research and critically revised the manuscript.

\bibliographystyle{IEEEtran}
\bibliography{mybib}

\end{document}